\pdfoutput=1

\documentclass[11pt]{article}

\usepackage{ACL2023}

\usepackage{times}
\usepackage{latexsym}
\usepackage{graphicx}
\usepackage{subcaption}
\usepackage{tabularx}

\usepackage[T1]{fontenc}

\usepackage[utf8]{inputenc}

\usepackage{microtype}

\usepackage{inconsolata}

\usepackage{booktabs}

\title{Exploratory Study into Relations between Cognitive Distortions and Emotional Appraisals}

\author{Navneet Agarwal \\
  Institute of Computer Science \\
  University of Tartu \\
  \texttt{navneet.agarwal@ut.ee} \\\And
  Kairit Sirts \\
  Institute of Computer Science \\
  University of Tartu \\
  \texttt{kairit.sirts@ut.ee} \\}

\begin{document}
\maketitle

\begin{abstract}
In recent years, there has been growing interest in studying cognitive distortions and emotional appraisals from both computational and psychological perspectives.
Despite considerable similarities between emotional reappraisal and cognitive reframing as emotion regulation techniques, these concepts have largely been examined in isolation.  
This research explores the relationship between cognitive distortions and emotional appraisal dimensions, examining their potential connections and relevance for future interdisciplinary studies.
Under this pretext, we conduct an exploratory computational study, aimed at investigating the relationship between cognitive distortion and emotional appraisals. 
We show that the patterns of statistically significant relationships between cognitive distortions and appraisal dimensions vary across different distortion categories, giving rise to distinct appraisal profiles for individual distortion classes.
Additionally, we analyze the impact of cognitive restructuring on appraisal dimensions, exemplifying the emotion regulation aspect of cognitive restructuring.

\end{abstract}

\section{Introduction}
Understanding the intricate relationship between cognition, emotion, and behavior has long been a central focus of neuroscience and cognitive science. The advent of artificial intelligence (AI) and recent advances in natural language processing (NLP) have enabled computational researchers to contribute to this field by developing models capable of analyzing individuals' mental and emotional states from textual data. Within this rapidly evolving domain, the automated extraction of cognitive patterns that shape emotions and behaviors has gained significant traction, bridging the gap between psychological theories and computational innovation.

Emotions are expressed through various modalities, including tone of voice, facial expressions, gestures, and language, particularly in written text.
This multifaceted expression of emotions has attracted the interest from NLP and computational researchers in recent years \cite{wang2022systematic,plaza-del-arco-etal-2024-emotion}.
While discrete emotional states such as anger, joy, and fear are deemed universal and thus form the basis for automated emotion recognition research, a smaller number of studies have explored dimensional models, representing discrete emotions in continuous spaces \cite{plaza-del-arco-etal-2024-emotion}.
Appraisal theories
define emotions as responses that arise from an individual's evaluation of and event's significance to their personal goals and well-being, emphasizing that the quality and intensity of emotional responses depend on appraisals, which are the subjective interpretations of the situation \cite{moors2013appraisal}. In contrast to discrete emotional categories, appraisal theories maps an individual's emotional state to a continuous space with each dimension representing an appraisal dimension. This not only provides a more detailed understanding of a person's state, but also allows comparison between emotions.

Negative thoughts are a natural part of human experience; however, they can have a more profound impact on individuals with mental disorders, often becoming entrenched, automatic, and emotionally triggering. 
Cognitive distortions refer to irrationally exaggerated negative assessments of oneself or situations \cite{beck1963thinking} and they are linked to the states of depression \citep{joormann2016examining} and anxiety \citep{yazici2022interpersonal}. Moreover, cognitive distortions have been found to correlate with the use of non-adaptive emotion regulation strategies \citep{deperrois2022links}.
Cognitive restructuring, also known as cognitive reframing, is a therapeutic intervention designed to encourage a more positive outlook towards situations by addressing these negative thought patterns \cite{clark2013cognitive}. This technique involves replacing negative thoughts with more neutral or hopeful ``reframed thoughts'', which provide a softer alternative perspective on the situation. 



Although cognitive reappraisal and cognitive restructuring focus on different aspects of cognition--- reappraisal aims to change the appraisal of specific events, while restructuring addresses broader thinking patterns---they are both emotion regulation methods that target thoughts to influence emotional states.
This suggests the potential for a systematic relationship between emotional appraisal dimensions and cognitive distortions.
However, to our knowledge, this relationship has not been thoroughly explored. NLP offers a more accessible and efficient means for conducting such exploratory research compared to traditional psychological studies, which requires recruiting human subjects and eliciting relevant information from them. 

Our aim in this work is to bring together emotion appraisals and cognitive distortions, to explore the link between these two different but related psychological constructs. We believe that such a relation (if it exists) could be exploited to define more robust systems for automated emotion regulation.
For instance, understanding of such relations could help to devise more deliberate and personalized ways for encouraging  cognitive reframing and/or emotional reappraisal.
We begin by training appraisal prediction models to perform automated appraisal annotations on a dataset annotated with cognitive distortions
enabling a combined analysis of both constructs. We analyze the distribution of appraisal values for each distortion-appraisal pair individually, and find statistically significant relations between cognitive distortions and appraisal dimensions, suggesting that different distortion patterns may exhibit distinct appraisal profiles. 
Finally, when comparing the appraisal profiles between original and reframed texts, we observed a considerable positive shift in several appraisal dimensions, further demonstrating the link between the two constructs and supporting the need for combined study of the two areas. 


\section{Appraisal Modeling}
\begin{figure*}
    \centering
    \includegraphics[width=\linewidth]{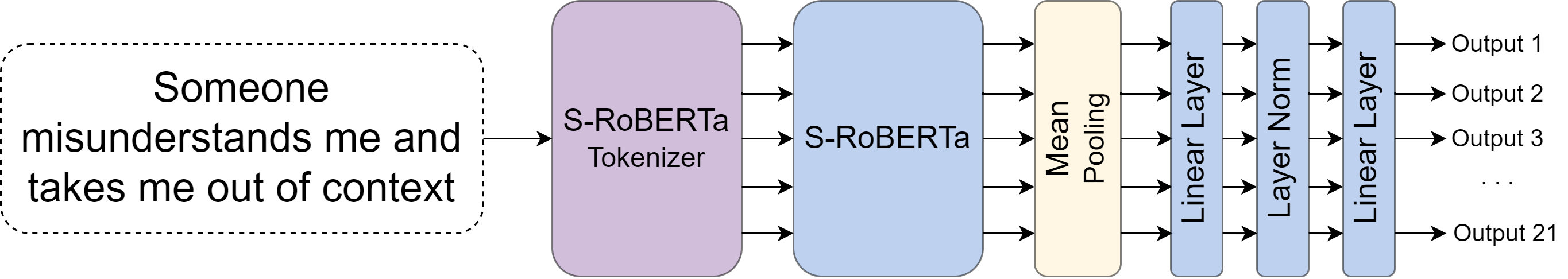}
    \caption{Overview of the appraisal prediction model.}
    \label{fig:model_def}
\end{figure*}

\label{sec:prediction_model}
In our attempt to analyze the relationship between emotional appraisals and cognitive distortions, we require both appraisal and distortion labels for the same text inputs. No such dataset with both labels is currently available. Therefore, we elected to perform automated data annotation in order to generate the desired labels. Although there are datasets that have been annotated for appraisals \cite{troiano-etal-2023-dimensional}, these are collected from neutral sources (since cognitive appraisal is a normative phenomenon of all emotions, functional and dysfunctional), and thus these texts are not likely to  contain too many cognitive distortions. To verify this, we conducted a preliminary experiment and applied a trained cognitive distortion prediction model to the appraisal-annnotated dataset \cite{troiano-etal-2023-dimensional}, which resulted in approximately 80\% data points assigned to the ``no distortion'' class, confirming our assumptions. Consequently, the alternate approach was adopted. In particular, we train appraisal prediction model on the \textit{crowd-enVent dataset} \cite{troiano-etal-2023-dimensional}, and apply it to the \textit{thinking trap dataset} \cite{sharma-etal-2023-cognitive}. The remainder of this section explains our methodology for training the appraisal prediction model.

\subsection{Crowd-enVent Dataset}
The \textit{crowd-enVent} is an emotion and appraisal based corpus of event descriptions collected by \citet{troiano-etal-2023-dimensional} as part of their research on emotional appraisals. During the data collection process, annotators recalled personal events and annotated them based on their recollection of emotions and feelings they experienced at the time of the event. The dataset contains 6600 event descriptions annotated with 21 appraisal dimensions on a 5-point Likert scale.\footnote{Appendix \ref{appendix:dimension_definitions} provides definitions of appraisal dimensions considered in this study.} The dataset is available in pre-defined splits of training (4320 entries), validation (1080 entries) and test (1200 entries) sets.\footnote{Available from \url{https://www.romanklinger.de/data-sets/crowd-enVent2023.zip}.} Please refer to the original paper by \citet{troiano-etal-2023-dimensional} for more details on the dataset.

\subsection{Model Architecture}

We use a multi-regression model  to predict the ratings of all appraisal dimensions simultaneously. Specifically, we adopt the multi-regression model  by \citet{milintsevich2023towards} who used it for predicting the severity of eight depression symptoms. 
The original model was a hierarchical model 
implemented with sentence-transformers to encode longer documents.
For our sentence-level prediction task, we forgo of the hierarchical definition of the model and directly use the sentence-level embeddings for final predictions. Furthermore, the prediction head now produces 21 regression outputs, one for each emotional appraisal dimension considered. 
Figure \ref{fig:model_def} provides an overview of the proposed model.

\subsection{Experimental Setup}
We use the S-RoBERTa Base model for encoding the input text, which is combined with the corresponding S-RoBERTa tokenizer\footnote{https://huggingface.co}. The model is trained using \textit{AdamW} optimizer with a learning rate of $10^{-5}$ and \textit{SmoothL1Loss} as the loss function. The network architecture applies a dropout of 0.3, along with layer norm regularization in the regression head. The overall code is a modified version of work by \citet{milintsevich2023towards}.\footnote{\url{https://git.unicaen.fr/kirill.milintsevich/hierarchical-depression-symptom-classifier}}

\subsection{Results}
\begin{figure}[h]
    \centering
    \includegraphics[width=\linewidth]{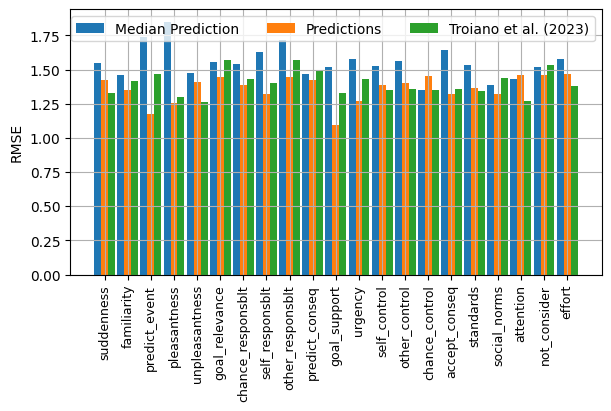}
    \caption{Appraisal ratings prediction accuracy in RMSE for our model, the results reported by \citet{troiano-etal-2023-dimensional}, and those calculated against median predictions for each appraisal dimension as baseline.}
    \label{fig:model_performance}
\end{figure}

Our model performs on par with the results from \citet{troiano-etal-2023-dimensional}. 
We use the root-mean-squared error (RMSE) as the evaluation metric and report macro-RMSE averaged over all appraisal dimensions on the test set of 1.36  compared to 1.40 reported in the original paper.
Furthermore, Figure \ref{fig:model_performance} compares the performance of the two models for each appraisal dimension, with our model out-performing \citet{troiano-etal-2023-dimensional} for 13 out of 21 appraisal dimensions. Figure \ref{fig:model_performance} also reports the RMSE of just predicting the median ratings for individual appraisal dimensions, with both trained models performing better than the baseline median predictor in most dimensions. The average RMSE of the median predictor was 1.55, which is clearly worse than the trained models. 
Thus, we deem our trained model good enough to be used for automated appraisal annotation in the remainder of the paper.

\section{Cognitive Distortion and Emotional Appraisal}
\label{sec:correlation}

With the aim of analyzing the relationship between appraisals and distortions, we apply the appraisal prediction model trained in the previous section to the \textit{Thinking Trap dataset} \cite{sharma-etal-2023-cognitive}, obtaining a dataset with 22 labels per input text (one cognitive distortion label and 21 emotional appraisal ratings). The resulting dataset forms the basis for the analysis discussed in the remainder of the paper. 
This section provides details on the dataset and explains the statistical methods used to analyze the relationships between cognitive distortions and appraisal dimensions.


\subsection{Thinking trap Dataset}

The \textit{Thinking Trap} dataset was collected by \citet{sharma-etal-2023-cognitive} as part of cognitive reframing research. The dataset contains entries from the existing \textit{Thought Records Dataset } \cite{burger2021natural}, along with additional entries collected through an online survey on the Mental Health America website, 
constituting 300 entries in total. This data was further annotated by mental health practitioners and clinical psychology graduate students, providing annotations for 15
cognitive distortion labels (14 distortions and one ``no distortion'' class)
addressed in the text along with corresponding reframed thoughts\footnote{Please refer to appendix \ref{appendix:dimension_definitions} for distortion definitions.}.

The original annotation process of the \textit{Thinking Trap} dataset resulted in a multi-label dataset, with each text associated with one or more distortion labels. Because in this research we are interested in each cognitive distortion category separately,
we converted the dataset into a multi-class format by repeating the data points once for each associated distortion label. The resulting dataset contained only 19 data points belonging to the ``no distortion'' class, amounting to only 1.9\% of the total data. Because we wanted to contrast appraisal profiles for texts with and without cognitive distortions, we included 77 data points with the ``no distortion'' class from an additional dataset also collected by the same authors.
The final dataset contains 1036 data points with the class distribution provided in Table~ \ref{tab:thinking_trap_distribution}. For more details on the dataset, please refer to the original work by \citet{sharma-etal-2023-cognitive}.

\begin{table}[t]
    \centering
    \begin{tabular}{lc}
        \toprule
        \textbf{Distortion Class} & \textbf{\# entries} \\
        \midrule
        All-or-nothing thinking & 99 \\
        Blaming & 34 \\
        Catastrophizing & 68 \\
        Comparing and Despairing & 12 \\
        Disqualifying the Positive & 40 \\
        Emotional reasoning & 43 \\
        Fortune telling	 & 78 \\
        Labeling & 102 \\
        Magnification & 15 \\
        Mind reading & 71 \\
        Negative feeling or emotion	 & 151 \\
        
        Overgeneralization & 107 \\
        Personalization & 98 \\
        Should statements & 22 \\
        \midrule
        Not distorted & 96 \\
        \bottomrule
    \end{tabular}
    \caption{Statistics of the cognitive distortion labels  of our version of the Thinking Trap dataset.}
    \label{tab:thinking_trap_distribution}
\end{table}

\subsection{Statistical Analysis}
To investigate the relationship between cognitive distortions and appraisal, we analyzed the statistical significance between each distortion category and each appraisal dimension. For each distortion-appraisal pair, we formed two groups of texts: a \textit{positive group (p)}, consisting of texts annotated with the cognitive distortion, and a \textit{negative group (n)}, consisting of texts without the distortion. This grouping allowed us to compare appraisal values in the presence and absence of each cognitive distortion. 

We performed an independent statistical analysis for each distortion-appraisal pair to isolate the effect of each distortion on the appraisal dimensions.
Specifically, we employed the non-parametric Mann-Whitney U Rank Test \cite{MannWhitney1947} to assess differences between the positive and negative groups. Under the null hypothesis, we posited that there would be no difference in appraisal values between the two groups (\textit{p}=\textit{n}, where \textit{p} represents the positive group and \textit{n} represents the negative group). To account for multiple comparisons across 14 cognitive distortion classes and 21 appraisal dimensions, we applied a Bonferroni correction \citep{abdi2007bonferroni}, setting a base p-value of 0.05, which was divided by the number of comparisons, which is 307 (the product of 14 and 21).


\begin{figure*}[!th]
    \centering
    \begin{subfigure}[b]{0.54\textwidth}
        \centering
        \includegraphics[width=\linewidth]{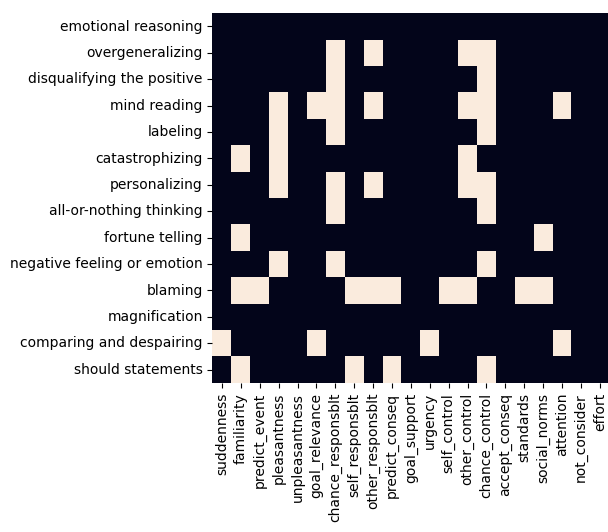}
        \caption{No-distortion}
    \end{subfigure}
    \hfill
    \begin{subfigure}[b]{0.43\textwidth}
        \includegraphics[width=\linewidth]{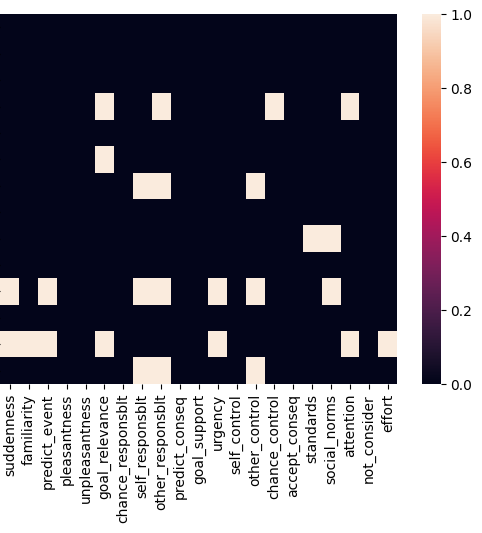}
        \caption{Exclusive}
    \end{subfigure}
    \caption{Discretized significance plots of distortion-appraisal pairs for two definitions of the negative groups. White cell implies a statistically significant relation between the distortion-appraisal pair, the black cell represents no statistically significant difference.}
    \label{fig:p_values}
\end{figure*}

\subsection{Negative groups}
One major consideration is the definition of the negative group within this analysis. The relationship between cognitive distortions and emotional appraisal dimensions, as inferred from the Mann-Whitney test, is strongly influenced by how the positive and negative groups are defined. Although the definition of the positive group \textit{p} is fixed, the negative group can have different meanings. Therefore, we consider the following three different definitions of the negative group \textit{n}:

\vspace{0.2cm}
\noindent \textbf{No distortion}: in this case, the negative group only contains entries without any distortion, i.e., those belonging to the ``no distortion'' class. This group represents the appraisal profile of texts without cognitive distortions and acts as a global baseline against which we can compare individual distortion profiles.

\vspace{0.2cm}
\noindent \textbf{Exclusive}: here, the negative group contains entries that do not belong to the given distortion class (defining the positive group) but belong to other cognitive distortion classes (excluding ``no distortion''). By utilizing this negative group, we can identify differences in appraisal values between various distortion classes. This approach enables us to analyze the appraisal profile of a specific distortion in relation to other distortions, rather than comparing it to a global baseline.

\vspace{0.2cm}
\noindent \textbf{All others}: in this configuration we combine both settings by including in the negative group all entries that do not belong to the specified distortion class (defining the positive group). 

The remainder of the paper mostly focuses on the first two categories of negative groups since behavior of \textit{all others} and \textit{exclusive} categories was found to be identical in all our experiments. This can be attributed to the fact that roughly 90\% of the data points express a distortion, thereby dominating the behavior of the ``no distortion'' class.


\subsection{Results}
Figure \ref{fig:p_values} plots the discretized significance values for 
two definitions of the negative groups
considered in our study.\footnote{Please refer to Appendix \ref{appendix:p_value_analysis} for the remaining plots.} In these plots, a cell colored white represents a statistically significant difference in appraisal values between the positive and negative groups, while a black cell indicates no statistically significant difference. 

First, we observe notable similarities between the two plots. For instance, both plots indicate a lack of statistical significance for \textit{emotional reasoning} and \textit{magnification} across all appraisal dimensions considered. Similarly, \textit{unpleasantness}, \textit{goal support}, and \textit{not consider} dimensions exhibit a lack of statistical significance across all distortion classes. However, we also observe certain differences between the two plots. 
In the ``no distortion'' setting (Figure~\ref{fig:p_values}(a)), appraisal dimensions like \textit{chance responsibility} and \textit{chance control} show a significant correlation with more than half of the cognitive distortions. However, in the ``exclusive'' setting (Figure \ref{fig:p_values}(b)), these dimensions lack significant correlation with any of the distortions.
While these plots reveal significant correlations between cognitive distortions and appraisal dimensions, they do not indicate the direction of strength of these correlations, thus motivating the further analysis conducted in the next section.

\section{Distortions and Corresponding Appraisal Profiles}
The previous section showed systematic relations between cognitive distortions and emotional appraisals. In this section, we delve deeper into these correlations, examining their nature, and studying specific appraisal profiles associated with each distortion class.

\begin{figure}
    \centering
    \includegraphics[width=\linewidth]{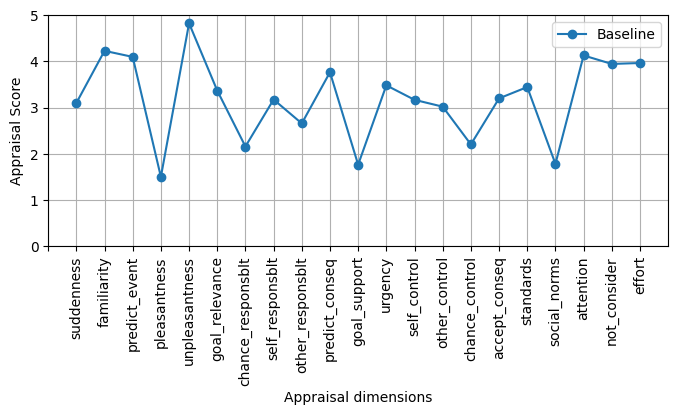}
    \caption{Baseline appraisal profile associated with the ``no distortion'' class.}
    \label{fig:baseline}
\end{figure}

\begin{figure*}
    \centering
    \includegraphics[width=0.9\linewidth]{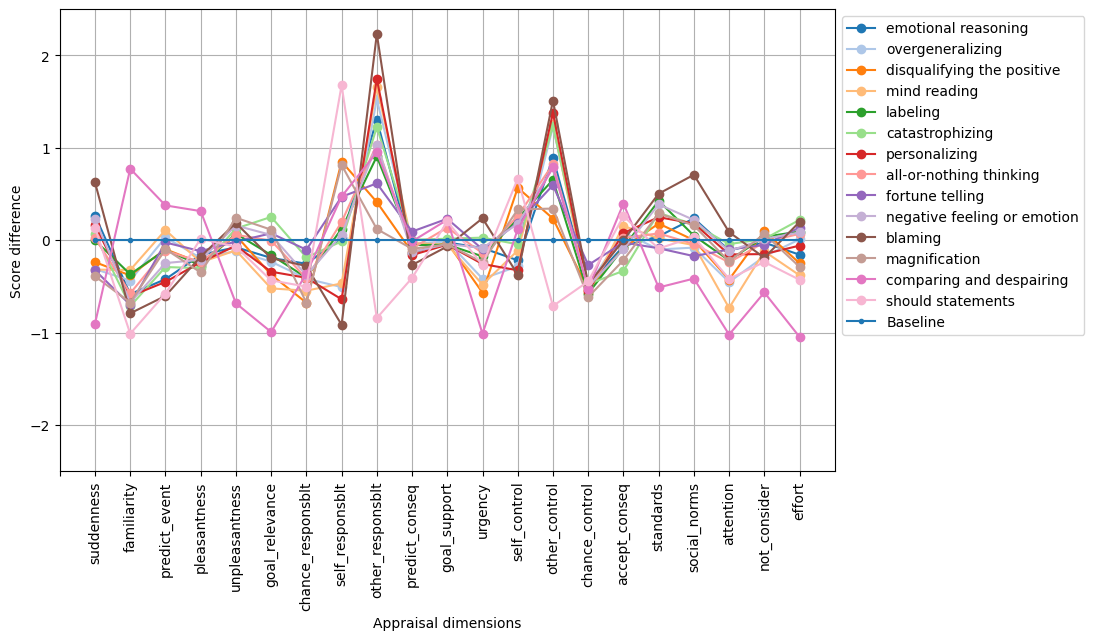}
    \caption{Appraisal profiles (relative to the ``no distortion'' baseline) for all distortion classes. The x-axis represents different appraisal dimensions considered while the y-axis plots the difference in appraisal scores between individual distortions and the baseline ($score(distortion) - score(baseline)$).}
    \label{fig:profile_plot}
\end{figure*}

\subsection{Methodology}
\label{sec:profile_definition}
We begin by defining the ``baseline'' appraisal profile using the ``no distortion'' negative group. 
This choice is motivated by the desire to establish a common baseline that represents the appraisal profile of inputs devoid of any cognitive distortion.
Specifically, the baseline profile is determined by calculating the median value for each appraisal dimension using the  data from the negative group. 
Similarly, the appraisal profile for each distortion class is defined by calculating the median value for each appraisal dimension using the corresponding positive group data.
Finally, to compare the distortion-specific appraisal profiles with the baseline "no distortion" profile, we subtract the baseline profile from each distortion-specific profile. In this manner, we obtain an appraisal profile for each cognitive distortion that is relative to the baseline profile.

\subsection{Results}
The baseline profile shown in Figure~\ref{fig:baseline} represents the median appraisal values of event descriptions without any cognitive distortions. Firstly, for most appraisal dimensions the median values cluster around the middle of the scale (between scores 2 and 4). Only four dimensions----\textit{familiarity}, \textit{predictability}, \textit{unpleasantness}, and \textit{attention}---exhibit median values above 4, while three dimensions---\textit{pleasantness}, \textit{goal support}, and \textit{social norms}---have median values below 2. We also notice a high median value for \textit{unpleasantness} stemming from the bias in the \textit{Thinking Trap} dataset, which primarily focused on negative thoughts and situations. 

Figure~\ref{fig:profile_plot} illustrates appraisal profiles associated with individual distortion classes, relative to the baseline profile. 
The distortion profiles generally exhibit similar patterns, compared to the baseline profile.
The two notable exceptions from others are \textit{should statements} and \textit{comparing and despairing}, which display deviations in the dimensions of \textit{familiarity, other's responsibility, and other's control}.
Regardless of the similarity of the overall pattern, some cognitive distortions show notable peaks in some appraisal dimensions, such as high \textit{self responsibility} for the \textit{should statements} or high \textit{other's responsibility} and low \textit{self responsibility} for \textit{blaming}.

Note that these plots illustrate the relative differences between the appraisal profiles of cognitive distortions and the baseline profile, but they do not indicate which of the distortion-appraisal relations were statistically significant. The following subsection discusses the appraisal profiles, considering the statistical significance analyses presented in Section~\ref{sec:correlation}.


\subsection{Discussion}
While Figure \ref{fig:p_values} reveals variations in the statistical significance of appraisal dimension correlations across different settings (presence/absence of distortions), Figure \ref{fig:profile_plot} demonstrates that the magnitude and direction of appraisal shifts relative to a non-distorted baseline are broadly similar across most distortion classes. This indicates that the presence of cognitive distortions, rather than their specific type, may be the primary driver of altered emotional experiences.


Furthermore, we observe that the appraisal dimensions of \textit{suddenness}, \textit{unpleasantness}, \textit{goal support}, \textit{accept consequences}, \textit{not consider}, and \textit{effort} exhibit a lack of significant correlation with nearly all distortion classes, as illustrated in Figure \ref{fig:p_values}(a). In Figure \ref{fig:profile_plot}, these dimensions also demonstrate a relatively balanced distribution of distortion profiles around the baseline. In contrast, the appraisal dimensions of \textit{chance responsibility}, \textit{other's responsibility}, \textit{other's control}, and \textit{chance control} show the highest number of significant correlations with distortion classes, as indicated in Figure \ref{fig:p_values}. Additionally, these dimensions exhibit highly polarized values in their distortion profiles, as seen in Figure \ref{fig:profile_plot}. This correlation between statistical significance and profile polarization suggests that most cognitive distortions exert similar effects on some appraisal dimensions.

\begin{figure}
    \centering
    \includegraphics[width=0.9\linewidth]{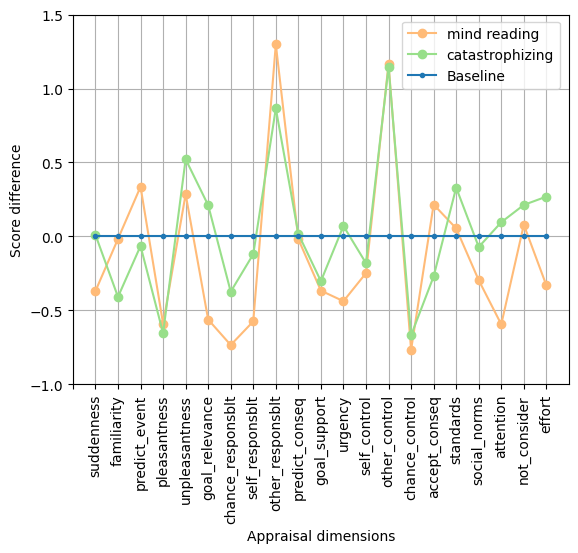}
    \caption{Appraisal profiles (relative to the baseline) for selected distortion classes. The x-axis represents different appraisal dimensions considered while the y-axis plots the difference in appraisal scores between individual distortions and the baseline ($score(distortion) - score(baseline)$).}
    \label{fig:profile_specific}
\end{figure}

Finally, we illustrate specific distortion-appraisal correlations from Figure \ref{fig:p_values} that align with established psychological principles. To this end, Figure \ref{fig:profile_specific} depicts the appraisal profiles for two distortion classes: \textit{mind reading} and \textit{catastrophizing}. In the case of \textit{mind reading}, the observed appraisal values for responsibility (namely, \textit{self responsibility}, \textit{other's responsibility}, and \textit{chance responsibility}) and control (namely, \textit{self control}, \textit{other's control}, and \textit{chance control}) are consistent with the distortion's underlying mechanism. \textit{Mind reading}, by definition, involves assuming knowledge of another person's thoughts and intentions, thereby implicitly attributing greater responsibility and control to that other person rather than to oneself or to chance. Conversely, \textit{catastrophizing} exhibits a negative correlation with the \textit{familiarity} and \textit{accept consequences} dimensions. This is psychologically plausible, as an individual's lack of familiarity with a situation would likely amplify feelings of uncertainty and uncontrollability, making the potential outcomes seem more catastrophic. Furthermore, a reduced ability to accept consequences would logically exacerbate the perceived severity of potential negative outcomes, thus fueling catastrophic thinking.

\section{Cognitive Reframing and Emotional Regulation}
\begin{figure*}
    \centering
    \includegraphics[width=0.9\linewidth]{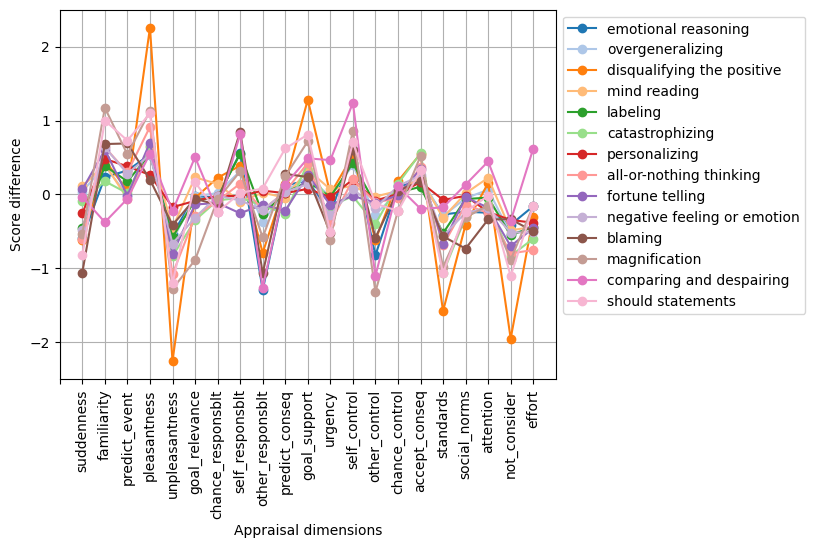}
    \caption{The shift in appraisal profiles after cognitive reframing. Positive values mean that median appraisal value increased after the reframing, while negative values mean indicate a decrease.}
    \label{fig:profile_plot_after}
\end{figure*}

In the final analysis of this research, we examine the impact of cognitive restructuring on appraisal profiles. Cognitive restructuring aims to regulate an individual's emotional state. Therefore, a significant change in appraisal profiles is expected following the restructuring process. Specifically, we anticipate a positive shift in the appraisal profiles as a result of cognitive restructuring.


\subsection{Dataset and Methodology}
The \textit{Thinking Trap} dataset, which formed the basis of our analysis thus far, also provides reframings of the input text. 
Annotators were asked to write reframes that are rational, specific, readable and actionable. 
Appraisal labels were generated for the reframes using automated annotation, employing the same prediction model trained in Section~\ref{sec:prediction_model}. 
Then, based on the reframed inputs, appraisal profiles for different distortion classes were generated using the same methodology detailed in Section~\ref{sec:profile_definition}. 


\subsection{Inferences and Interpretations}
Figure~\ref{fig:profile_plot_after} plots the difference between appraisal profiles generated from the reframed inputs and the original profiles shown in Figure~\ref{fig:profile_plot}. The plots illustrate a considerable increase in values for \textit{pleasantness}, while showcasing a decrease in values for dimensions such as \textit{unpleasantness} and \textit{not consider}. The observed increase in \textit{pleasantness}, along with the decrease in \textit{unpleasantness}, indicates a positive shift in people's perception of the situations. We also observe a decrease in the appraisal dimension \textit{not consider} across all distortion classes, reflecting an increased willingness to engage with, rather than avoid, situations. 
Overall, the changes in appraisal dimensions plotted in Figure~ \ref{fig:profile_plot_after} provide evidence that cognitive reframing is associated with a potential positive shift in emotional appraisal.
This finding further strengthens the proposed link between cognitive distortions and emotional appraisals, and the necessity of studying these concepts together.
Finally, this positive shift in appraisal profiles for reframed texts also supports the validity of the automated appraisal prediction model used in our study.


\section{Related Work}

Although automatic prediction of emotion appraisals has been less studied than predicting discrete emotions, in recent years, several works have emerged in this direction. Several papers have contributed datasets annotated with different sets of appraisal dimensions \cite{hofmann2020appraisal,troiano2022x,troiano-etal-2023-dimensional,zhan2023evaluating}. Few experiments have been presented to demonstrate the utility of adopting NLP models to predict the appraisal values based on text, training CNN-based neural classifier \cite{hofmann2020appraisal}, fine-tuning RoBERTa-based models \cite{wegge2022experiencer,troiano-etal-2023-dimensional}, or prompting large language models (LLMs) \cite{zhan2023evaluating}.

In relation to cognitive distortions, similarly, few datasets annotated with cognitive distortion categories have been published \cite{shreevastava-foltz-2021-detecting,wang-etal-2023-c2d2,sharma-etal-2023-cognitive} with various classification approaches adopted to predict the cognitive distortions from text, using both fine-tuned BERT-based models \cite{tauscher2023automated,maddela-etal-2023-training} and recently also increasingly prompting LLMs \cite{chen-etal-2023-empowering,lim2024erd}.
Furthermore, NLP researchers have developed methods for a variety of reframing tasks including sentiment and empathy writing \cite{reif2022recipe, sharma2023human}, positive reframing \cite{ziems2022inducing,goel2024socratic,jia-etal-2025-positive}, and cognitive restructuring \cite{sharma-etal-2023-cognitive,maddela-etal-2023-training,zhan2024large,xiao2024healme}.

\section{Conclusion}
Since both cognitive restructuring and emotional reappraisal serve as emotion regulation strategies, this paper explored the connection between the two constructs from a computational standpoint. 
As a first step, we automatically annotated appraisal ratings on a dataset of cognitive distortions, producing a new dataset supporting combined analysis.
Our analysis at the distortion-appraisal pair level revealed statistically significant relations between cognitive distortions and emotional appraisal dimensions, demonstrating systematic links between the two constructs. 
By constructing appraisal profiles for individual cognitive distortions and comparing them to a ``no distortion'' baseline, our analysis showed similar patterns across distortion profiles, indicating a clear distinction between profiles with and without cognitive distortions.
Analyzing the impact of cognitive restructuring on the appraisal profiles revealed a shift towards appraisal values indicative of a more positive interpretation of the situations, consistent with the established definition of cognitive reframing.
It is our hope that these preliminary results demonstrate the existence of a relationship between cognitive distortions and emotional appraisal dimensions, and illustrate the potential benefits of jointly studying these constructs and motivating further computational research in this area.

\section*{Limitations}

While we believe this research represents first steps toward understanding correlations between cognitive distortions and emotional appraisals, some concerns remain. A primary issue in computational mental health research is the quality of available data and annotations. The \textit{Thinking Trap} dataset also suffers from the same problem: manual examination reveals instances with incorrect distortion labels. 
This classification error stems, in part, from the subjective nature of the task, leading to inconsistencies in labeling even among experienced psychologists. Another concern with this dataset is the length of the input texts. Classifying cognitive distortions or any mental health related aspects based on such short texts is unrealistic and contributes to noise in the data. Some examples illustrating this issue are included in Appendix~\ref{appendix:data_limitations}.

Another major limitation of this work is that it is exploratory research conducted from a computational standpoint, which lacks the in-depth considerations and reasoning from a psychological perspective. This research needs to be complemented by detailed psychological studies, but this is beyond the scope of this paper as well as our knowledge and expertise.

\section*{Ethical Considerations}
Despite the growing use of NLP (and AI in general) for analyzing the mental and emotional state of individuals based on a variety of input data sources, some ethical considerations need to be taken into account within the research process.

One major area of consideration is the data collection process for such studies. In this work, we use an existing publicly available dataset whose authors reported considering the ethical aspects of their data collection and annotation and also sought approval for their procedures from their institution's review board (see \citep{sharma-etal-2023-cognitive} Section~4.3).

This research direction also comes with significant ethical considerations pertaining to the use of such models. Despite the growing interest in automated systems for mental health analysis and monitoring, improper use of these systems can lead to issues like labeling and stigma. Within our study, we are not developing systems for making predictions about an individual's mental health, but rather studying the general patterns over groups.

\bibliography{mybib}
\bibliographystyle{acl_natbib}

\newpage
\appendix
\section*{Appendix}
\section{Statistical analyses}
\label{appendix:p_value_analysis}

\begin{figure}[h]
    \centering
    \begin{subfigure}[b]{\linewidth}
        \centering
        \includegraphics[width=\linewidth]{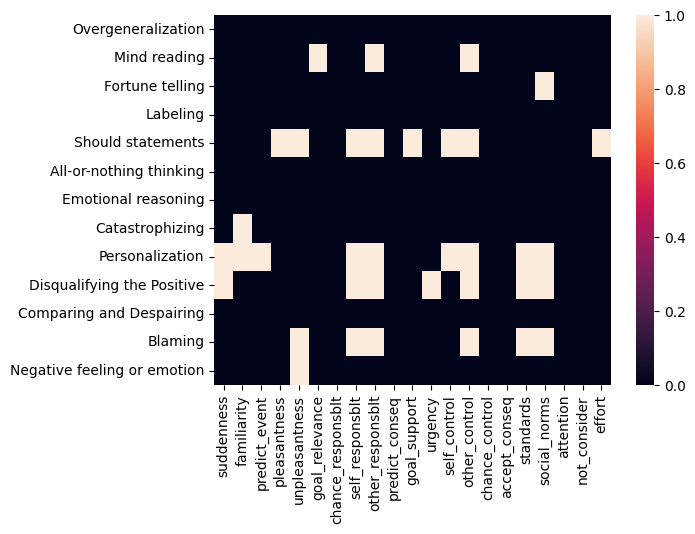}
        \caption{All others}
    \end{subfigure}
    \begin{subfigure}[b]{\linewidth}
        \centering
        \includegraphics[width=\linewidth]{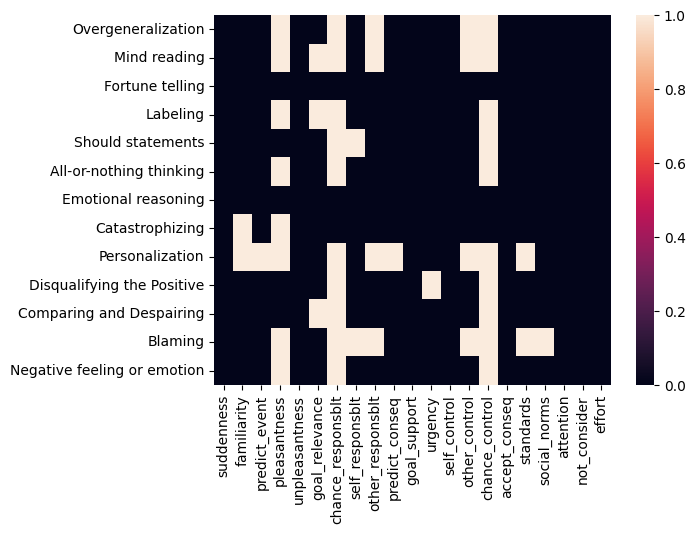}
        \caption{No-distortion}
    \end{subfigure}
    \begin{subfigure}[b]{\linewidth}
        \centering
        \includegraphics[width=\linewidth]{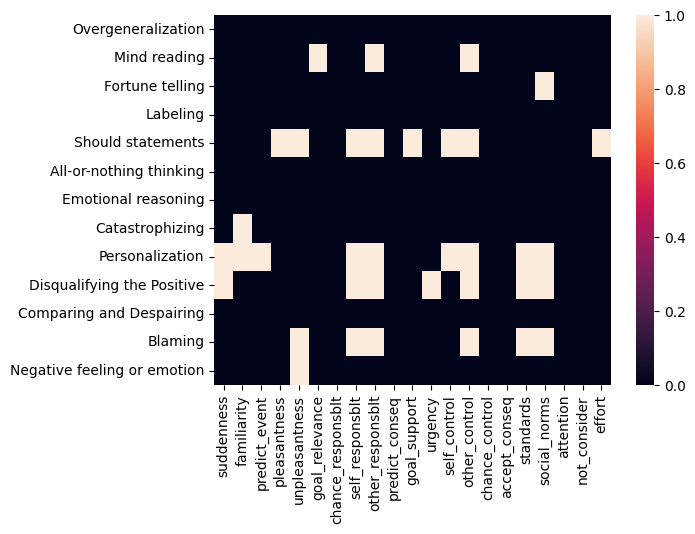}
        \caption{Exclusive}
    \end{subfigure}
    \caption{Significance plot between distortions and appraisals for different definitions of negative distribution}
    \label{appendix:fig:p_values}
\end{figure}

Since the \textit{Thinking Trap} dataset ensured a uniform distribution of data points across different distortion classes (Table~\ref{tab:thinking_trap_distribution}), the appraisal values of texts from cognitive distortion categories outweigh those from the ``no distortion'' class, resulting in identical plots for the ``Exclusive'' (Figure~\ref{appendix:fig:p_values}(c)) and ``All others'' (Figure~\ref{appendix:fig:p_values}(a)) settings.

\section{Data Limitations and Corresponding Examples}
\label{appendix:data_limitations}
The datasets used in this research also exhibit certain quality issues common to most datasets in the field. In the \textit{Thinking Trap }dataset, some input thoughts are extremely short, making it difficult to assess them due to insufficient information. Some examples of such cases are provided below:

\begin{itemize}
    \item I had a breakup, I am the cause of the breakup.
    \item I gained weight, I feel like I need to die to be happy.
    \item My diet is not working, I feel like a failure.
\end{itemize}

While these texts may provide some hints about potential cognitive distortions, both models and humans would struggle to accurately assess 21 different emotional appraisal dimensions.


\section{Label definitions}
\label{appendix:dimension_definitions}

\subsection{Emotion Appraisals}
See Table~\ref{appendix:tab:appraisal_dimensions} for the definitions of the 21 emotion appraisal dimensions used in this study.

\begin{table*}[h]
    \centering
    \begin{tabular}{|p{0.24\linewidth}|p{0.75\linewidth}|}
        \hline
        Dimension & Definition \\
        \hline
        \hline
        Suddenness & The event was sudden or abrupt \\
        Familiarity & The event was familiar \\
        Event predictability & I could have predicted the occurrence of the event \\
        Pleasantness & The event was pleasant \\
        Unpleasantness & The event was unpleasant \\
        Goal relevance & I expected the event to have important consequences for me \\
        Situational responsibility & The event was caused by chance, special circumstances, or natural forces \\
        Self responsibility & The event was caused by my own behavior \\
        Others responsibility & The event was caused by somebody else’s behavior \\
        Anticipated consequence & I anticipated the consequences of the event \\
        Goal support & I expected positive consequences for me \\
        Urgency & The event required an immediate response \\
        Self control & I was able to influence what was going on during the event \\
        Others control & Someone other than me was influencing what was going on \\
        Chance control & The situation was the result of outside influences of which nobody had control \\
        Consequence acceptance & I anticipated that I would easily live with the unavoidable consequences of the event \\
        Internal standards & The event clashed with my standards and ideals \\
        External standards & The actions that produced the event violated laws or socially accepted norms \\
        Attention & I had to pay attention to the situation \\
        Not consider & I tried to shut the situation out of my mind \\
        Effort & The situation required me a great deal of energy to deal with it \\
        \hline
    \end{tabular}
    \caption{List of appraisal dimensions considered in this research.}
    \label{appendix:tab:appraisal_dimensions}
\end{table*}

\subsection{Cognitive Distortions}
See Table~\ref{appendix:tab:cognitive_distortion} for the definitions of the 14 cognitive distortion categories used in this study.

\begin{table*}[h]
    \centering
    \begin{tabular}{|p{0.28\linewidth}|p{0.71\linewidth}|}
        \hline
        Dimension & Definition \\
        \hline
        \hline
        Emotional reasoning & Treating your feelings like facts. \\
        Overgeneralization & Jumping to conclusions based on one experience. \\
        Disqualifying the positive & When something good happens, you ignore it or think it does not count. \\
        Mind reading & Assuming that you know what someone else is thinking. \\
        Labeling & Defining a person based on one action or characterstic. \\
        Catastrophizing & Focusing on the worst/case scenario. \\
        Personalizing & Taking things personally, or making them about you\\
        All-or-nothing thinking & Thinking in extremes. \\
        Fortune telling & Trying to predict the future. Focusing on one possibility and ignoring the other, more likely outcome\\
        Negative feeling and emotion & Getting "stuck" on a distressing thought, emotion, or belief. \\
        Blaming & Giving away your own power to other people. \\
        Magnification & Exaggerating certain aspects of yourself, other people, or a situation while often simultaneously downplaying others. \\
        Comparing and despairing & Comparing your worst to someone else's best. \\
        Should statements & Setting unrealistic expectations of yourself. \\
        \hline
    \end{tabular}
    \caption{List of cognitive distortions considered in this research.}
    \label{appendix:tab:cognitive_distortion}
\end{table*}





\end{document}